\documentclass[journal,transmag]{IEEEtran}
\ifCLASSINFOpdf
\else
\fi
\usepackage[inline]{enumitem}

\usepackage{booktabs}
\usepackage{subcaption}
\usepackage{multicol}

\usepackage{amssymb}
\usepackage{amsmath}

\usepackage{graphicx}
\usepackage{cleveref}

\begin{document}
%


\title{Representation learning of rare temporal conditions\\for travel time prediction}

%

\author{\IEEEauthorblockN{Niklas Christoffer Petersen\IEEEauthorrefmark{1},
Filipe Rodrigues\IEEEauthorrefmark{2}, and
Francisco Camara Pereira\IEEEauthorrefmark{2}, 
\IEEEauthorblockA{\IEEEauthorrefmark{1}Movia Public Transport, 
    Gammel Køge Landevej 3, 
    Valby, Denmark}}
\IEEEauthorblockA{\IEEEauthorrefmark{2}Department of Management Engineering, 
   Technical University of Denmark,
   Kgs. Lyngby, Denmark}
\thanks{Corresponding author: F. Pereira (email: camara@dtu.dk)}}

\markboth{}%
{}
%



\IEEEtitleabstractindextext{%
\begin{abstract}
Predicting travel time under rare temporal conditions (e.g.\ public holidays, school vacation period, etc.) constitutes a challenge due to the limitation of historic data. If at all available, historical data often form a heterogeneous time series due to high probability of other changes over long periods of time (e.g.\ road works, introduced traffic calming initiatives, etc.). This is especially prominent in cities and suburban areas. We present a vector-space model for encoding rare temporal conditions, that allows coherent representation learning across different temporal conditions. We show increased performance for travel time prediction over different baselines when utilizing the vector-space encoding for representing the temporal setting.
\end{abstract}

}

\maketitle

\IEEEdisplaynontitleabstractindextext

%
\IEEEpeerreviewmaketitle


\section{Introduction}
\label{sec:intro}

\begin{figure*}[ht!]
    \centering
    \begin{subfigure}[b]{0.75\textwidth}
         \centering
         \includegraphics[width=\textwidth]{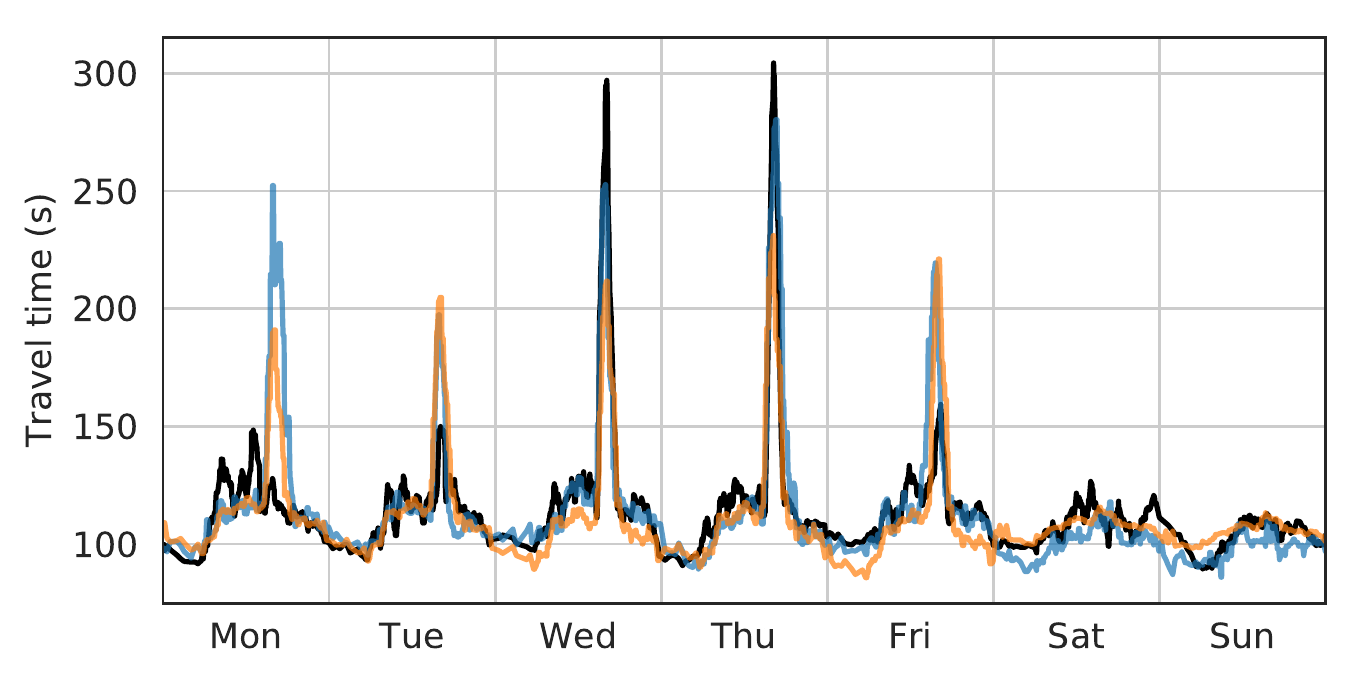}
         \caption{Example of \textit{normal} week without any rare temporal conditions, Week 15, 2019. True travel time follows to a large extend the expected weekly cyclic patterns (Baseline).}
         \label{fig:weekly_pattern_normal}
     \end{subfigure}
     \hfill
     \begin{subfigure}[b]{0.75\textwidth}
         \centering
         \includegraphics[width=\textwidth]{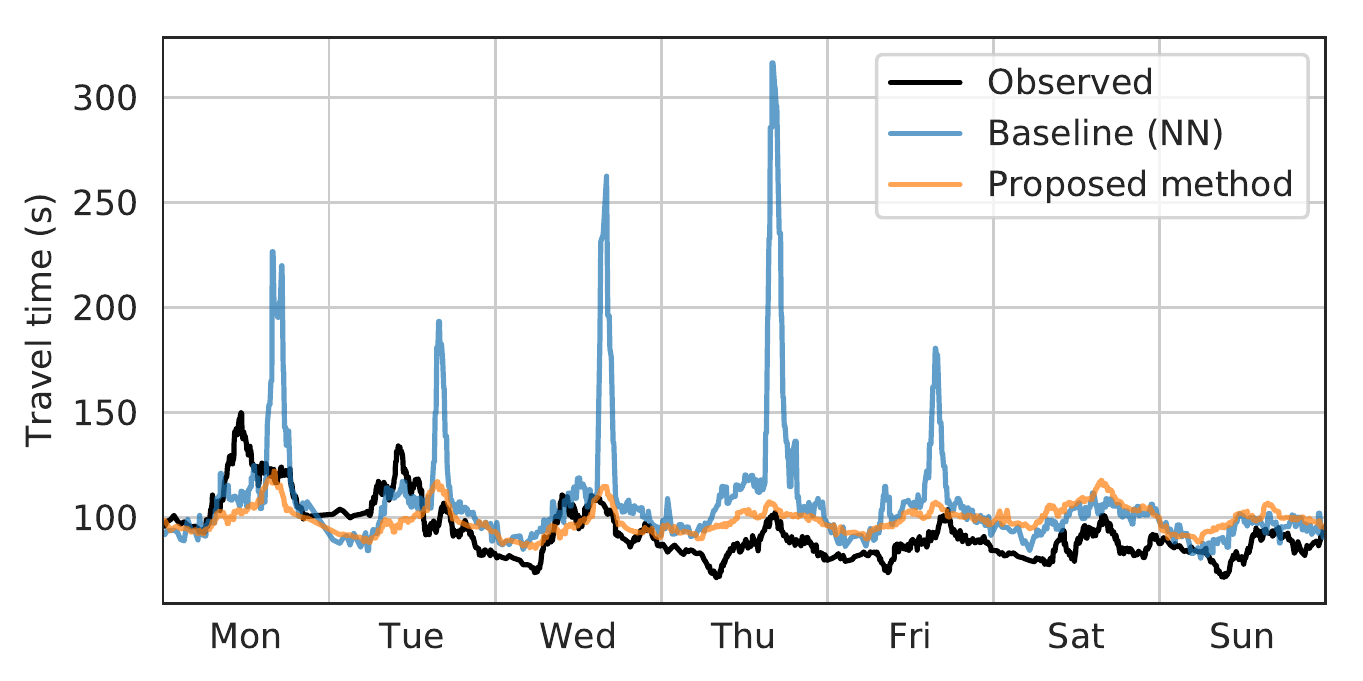}
         \caption{Example of \textit{special} week, the week of Easter, Week 16, 2019. True travel time deviates greatly from expected weekly cyclic patterns (Baseline).}
         \label{fig:weekly_pattern_special}
     \end{subfigure}
    \caption{Predicted and observed travel time for the same road segment for two consecutive weeks.}
    \label{fig:weekly_pattern}
\end{figure*}

Travel demand and travel time variability in suburban and urban areas are highly affected by relative short-phased cyclic patterns, e.g.\ systematic daily and weekly recurrent variations due to peak hours. The short-phased patterns often contribute significantly to the variation, especially in urban and congested areas as illustrated in Figure~\ref{fig:weekly_pattern_normal}, where the afternoon peek hour can be easily identified throughout the weekdays. However, a number of other factors are known to also affect travel demand and travel time, including weather, events, and more rare temporal conditions such as holidays and school vacations periods. \Cref{fig:weekly_pattern_special} is a evidence of this, showing the same road segment as (a) but the following week, which happens to be the week of Easter. We will later detail the baseline, which fails and greatly overestimate the travel times where the peak hours \textit{usually} are; and the proposed method using representation learning.

Traditionally, research within the transportation field mainly focus on achieving best performance for weekdays, especially peak hours, where congestion is worst and most people use the transport systems. Therefore it might not seem like a first priority to precisely model the more rare temporal conditions, but we still see several arguments for the value of our work: 
\begin{itemize}
    \item Even though we might consider holidays, school vacations and weekends as rarer than workdays, they actually constitute 47\% of the calendar year\footnote{Calculated based on the official Danish school vacation calendar and official public holidays in Denmark.}. If we do not consider weekends and school vacations the holidays periods still constitute 6\% of the calendar year\footnotemark[1].
    \item This method can also improve the prediction of weekdays, since a good model of holidays and school vacations periods will allow the reduction of spillover noise from these rarer temporal conditions for impacting predictions for weekdays.
\end{itemize}

It is though important to emphasize that such rare temporal conditions do not constitute a single distribution of travel demand, and thus travel time. Some holidays will entitle very little demand (e.g.\ Christmas Day), while others actually can be more busy travel days than an average weekday (e.g.\ Christmas Eves Day) in some regions and times of the day, because people are meeting with families, friends, etc.

We argue in order to capture the isolated impacts of special and rare temporal conditions such as Christmas, School vacation, etc. we need to observe at least some occurrences of such impacts. However, this also constitute a non-trivial challenge, since underlying structures often have changed in between the occurrences of rare events. Thus simply encoding the temporal condition na\"ivly, e.g. using one-hot representation, will be too limited for rare conditions, and often just capture any underlying structural changes that has occurred in between. Similarly, since the events are rare by definition, missing historical data is a significant issue. E.g. every time a new road segment is added to the (observed) road network, we will need to wait years to collect enough occurrences of each rare event. 


\subsection{The travel time prediction problem}
To demonstrate our methodology we apply it to the \textit{travel time problem}, specifically using data from bus trips in the Greater Copenhagen Area.



Most research of the travel time and bus travel time prediction problems does not at all consider changes to the road network or changes to location of stop points/intersections, etc. Even though they at the same time use long periods of historic data for estimating their models, thereby increasing the likelihood of changes having occurred that were not captured and understood by the model.


The main contributions of our work are as follows:
\begin{itemize}
    \item The identification and selection of \textit{representative time series} for link travel time prediction under different (rare) temporal conditions.
    \item A method for learning the coherent semantic structure of both rare and frequent temporal conditions from these representative time series using an embeddings model. 
    \item Experiments and the application of the above on the link travel time prediction problem, demonstrating the methods generalizable properties to links with insufficient historical data and/or links with structural changes impacting such historical data.
\end{itemize}

\subsection{Literature review}
Traditional, studies of predictive modeling approaches for travel time prediction focuses primarily on peek times, and to a smaller extent non-peek and weekends. E.g.\ \cite{Zhang2015, Lin2013, Wang2018} include day of week as features for travel time, but does not state any special care with respect to holidays or other rare temporal conditions. Likewise, \cite{Jeong2005, Ma2019} defines four time periods: \textit{weekend}, \textit{weekday peak}, \textit{weekday off-peak}, and \textit{weekday evening} as explanatory variables for predicting public transport travel times. Wai et al. \cite{Wai2020} present a system for bus travel time prediction, and includes three specific public holidays as explanatory variables (Christmas Day, Boxing Day, and Easter Monday). Several studies \cite{Ando2006, Hojati2016} specifically remove holidays from their data sets as part of their data preparation. We have only found very little research that focuses specifically on handling rare temporal conditions such as public holidays, school vacation periods, etc., even when extending our research to related fields:

Cools et al. \cite{Cools2010} present a travel survey, and uses \textit{no holiday}, \textit{holiday}, and \textit{summer holiday} as explanatory variables for travel behavior.
Barrow and Kourentzes \cite{CallCenterSpecialDays} use an ANN model for predicting call centre arrivals, and encodes special days using binary encoding, i.e.\ \textit{Holiday}/\textit{No holiday}.

In all these cases the representation of rare temporal conditions is either completely absent, or only to a small degree captured using either binary encoding or one-hot encoding \cite{Lin2013}.
 
A deep learning approach for car travel time is presented by Wang et al. \cite{Wang2018}, in which the use of embeddings are used to encode day of week.

\subsection{Embeddings}
In \textit{Natural Language Processing} (NLP) research, the concept of \textit{word embeddings} was initially pioneered by \cite{Embeddings2003,Embeddings2008} to represent similarity between words. Each word is \textit{embedded} into a real valued vector $\mathbf{w} \in \mathbb{R}^D$, where $D$ is dimensionality of the embedding vector space. The key is that the learned vector representation of words represents their semantical similarity. Recent research \cite{Word2Vec, Embeddings2014} have shown \textit{word embeddings} to generalize across various NLP tasks, and to be very successful as a initial tranformation step, outperforming other methods for many downstream NLP tasks.

The concept can be generalized to any symbolic data that is semantically distributed, e.g.\ temporal conditions. Initially we represent each temporal condition at time $t$, as a \textit{one-hot} encoded vector, $\mathbf{c}^\text{oh}_t \in \left\{0,1\right\}^C$, where $C$ is the number of different temporal conditions captured by the model, and $\sum \mathbf{c}^\text{oh}_t=1$ for any time, $t$. We transform this encoding of the condition into it's semantic vector space representation cf.~(\ref{eq:encode}) from the \textit{embeddings weights matrix}, $\mathbf{W}^\textit{emb} \in \mathbb{R}^{C \times D}$. In Section~\ref{sec:embeddings_model} we show our method for estimating $\mathbf{W}^\textit{Emb}$ in more detail.

\begin{equation}
    \text{Encode}(\mathbf{c}^\text{oh}_t) = \mathbf{c}^\text{oh}_t \mathbf{W}^\textit{emb}_i
    \label{eq:encode}
\end{equation}


\section{Data acquisition and preparation}
We briefly describe the data sets needed for applying our proposed methodology and experiments, including possible construction from raw GPS trajectories. We assume the two following data sets are available:
\begin{enumerate*}[label=(\roman*)]
    \item \textit{Link travel times}, and
    \item \textit{Calendar with temporal condition labeling}
\end{enumerate*}.

\begin{table}[!ht]
  \centering
  \footnotesize
  \begin{tabular}{llr}
  \toprule
    Timestamp & Link ref. & Link travel time (s) \\ \midrule
    2019-10-10 00:20:02 & 29848:1254 & 63 \\ \midrule
    2019-10-10 00:21:07 & 1254:1255  & 65 \\ \midrule
    2019-10-10 00:21:51 & 1255:10115 & 44 \\ \midrule
    \vdots & \vdots & \vdots \\ \bottomrule
  \end{tabular}
  \caption{Example of raw travel time measurements from AVL system.}
  \label{tab:data}
\end{table}

For (i) we expect link travel time measurements to be available in a tabular form as illustrated in~\Cref{tab:data}. Each link travel time measurement has a timestamp, and a reference to the link (e.g.\ road segment) and the observed travel time used to traverse the link at the observed timestamp. This output is standard for most traffic monitoring systems, AVL systems used in public transport systems, and can otherwise efficiently be constructed via map matching of raw vehicle GPS trajectory data \cite{Newson2009, Jagadeesh2016}, thus allowing the proposed method to generalize to other domains and input data formats.

For the calendar data set (ii) we assume a very simple structure as shown in the example in Table~\ref{tab:cal}, that essentially maps any timestamp, $t$, to $\textbf{c}^\text{oh}_t$. We also include a labeling indication if it is a rare condition or not.
\begin{table}[ht!]
    \centering
    \footnotesize
\begin{tabular}{lp{2.5cm}p{.9cm}p{2.2cm}}
\toprule
Date &                          Temporal condition label &  Rare\newline condition &           One hot\newline encoding, $\textbf{c}^\text{oh}_t$ \\
\midrule
2018-12-29 &  Between Christmas/New Year Days &               True &  [1, 0, 0, 0, $\ldots$, 0] \\ \midrule
2018-12-30 &  Between Christmas/New Year Days &               True &  [1, 0, 0, 0, $\ldots$, 0] \\ \midrule
2018-12-31 &               New Year Eve's Day &               True &  [0, 1, 0, 0, $\ldots$, 0] \\ \midrule
2019-01-01 &                   New Year's Day &               True &  [0, 0, 1, 0, $\ldots$, 0] \\ \midrule
2019-01-02 &                        Wednesday &              False &  [0, 0, 0, 1, $\ldots$, 0] \\ \midrule
2019-01-03 &                         Thursday &              False &  [0, 0, 0, 0, $\ldots$, 0] \\
\bottomrule
\end{tabular}
    \caption{Holiday calendar with day type labelling}
    \label{tab:cal}
\end{table}

This is just a labeling of each day in the desired time frame, where any special day is labeled accordingly from domain knowledge, and all non-special days (e.g.~normal days) are simply labelled with their respectively weekday name. Since special days varies for country to country, and possibly between regions within the same country this part must be adjusted to the spatial frame of the travel time data (i). Furthermore, as this labeling will determine the temporal condition for e.g.~travel time, it may be necessary to include special day labels for more than just official public holidays. For instance, in the example in Table~\ref{tab:cal}, we want the days between Christmas and New Year to be modelled with a distinct temporal condition even though this period is not considered as officially holidays. A special temporal condition does not necessarily need to be labelled for the entire duration of a day, although it makes the input data set (ii) easier to produce. 

\section{Methodology}
Our goal is to estimate $\mathbf{W}^\textit{emb}$, such that it represents the learned semantic distribution of the defined temporal conditions. This allows it to be used for future travel time predictions. The approach is, that within a large sample of link travel time measurements, there exists overall homogeneous links with \textit{representative time series}, e.g.\ links that has not been subject to significant changes in the underlining structures as discussed in Section~\ref{sec:intro}. Furthermore, the frequency of measurements for \textit{representative time series} must be high enough, such that with a minimal level of imputation, they will constitute a \textit{regular} time series. The last condition avoids the need for dealing with large gaps in training data, and to our experience enhance the learning performance.

From the link travel time measurements (Table~\ref{tab:data}) we calculate $\mathbf{M}^\textit{ln}$ for each link, $\textit{ln}$, where $\mathbf{M}^\textit{ln}_{i,j}$ is the mean link travel for the $j$'th discrete time interval (e.g.~hour) of the $i$'th day. For combinations of $i, j$ that contain no travel time observations for the link we assign $\mathbf{M}^\textit{ln}_{i,j} = \bot$. The reason to include discrete time intervals on an explicit axis is to be able to learn the difference between temporal conditions that affects differently over the course of a day. 

\subsection{Selecting representative time series}

We define $\textit{Cov}^\textit{ln} \in [0; 1]$ as the coverage of link, $\textit{ln}$, i.e.\ how sparse/frequent are the observations of that link, where $N$ is the number of indices in $\mathbf{M}^\textit{ln}$:
\begin{align}
\textit{Cov}^\textit{ln}= \frac{1}{N} \sum_{i, j} h(\mathbf{M}^\textit{ln}_{i,j})
&
\;\;\text{where}\;\;
h(x) = \left\{
\begin{array}{l}
                  1 \; \text{if} \; x \neq \bot \\
                  0 \; \text{if} \; x = \bot
                \end{array}
                \right.
\end{align}

Initially we prune all links with $\textit{Cov}^\textit{ln} < \beta$, since training the embeddings model on too sparse samples will yield poor results from our experience. Ensuring high coverage also allows  application of 2D gaussian \textit{KNN imputation} \cite{KnnImputation} with a reasonable small enough  kernel on $\mathbf{M}^\textit{ln}$, yielding a complete covered matrix~$\mathbf{\widehat{M}}^\textit{ln}$, i.e.\ without any indices having the value~$\bot$.

Finally we want the representative links to be overall stationary, since non-stationarity in the time series suggest underlying  structural changes. We apply the \textit{PELT (Pruned Exact Linear Time)} change point algoritm \cite{Truong2020} on $\mathbf{\widehat{M}}^\textit{ln}$ for links,~$\textit{ln}$, where $\textit{Cov}^\textit{ln} > \beta$. We use the number of identified regimes, $R^\textit{ln}$, to further prune the selection of representative links, i.e.\ $R^\textit{ln} \leq \gamma$. 

\subsection{Embeddings model}
\label{sec:embeddings_model}

\begin{figure}[!ht]
    \centering
    \includegraphics[scale=1.1]{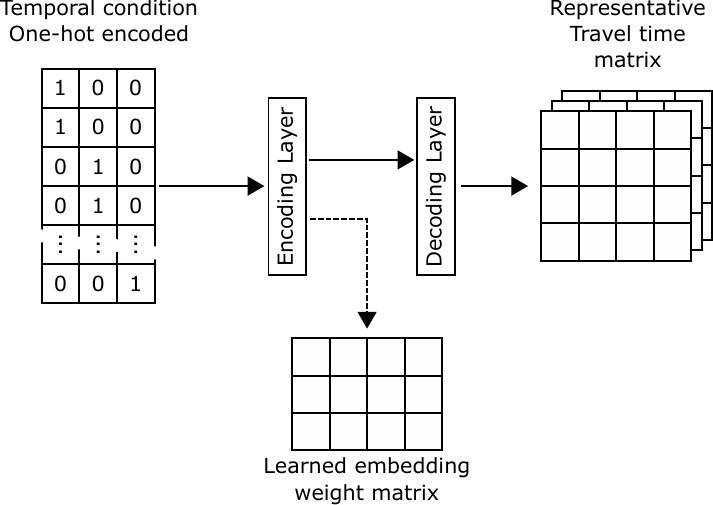}
    \caption{Embeddings model}
    \label{fig:embeddingsmodel}
\end{figure}

The selected representative links are used to train an embeddings model in order to learn $\mathbf{W}^\textit{emb}$ from $\mathbf{\widehat{M}}_t$ and temporal conditions, $\mathbf{c}^\text{oh}_t$. By both including rare and frequent temporal conditions in the same representation, e.g.\ specific holidays, respectively normal weekdays, we allow learning the general relation between the conditions. This will later allow our link travel time model to predict for rare conditions, even though no rare conditions has ever been observed for a particular link. 

The embeddings model is a simple encoder/decoder MLP as shown in Figure~\ref{fig:embeddingsmodel}. The embeddings model is consisting of just two linear layers: $L_1$: embedding of temporal condition, and $L_2$: fully connected linear layer to predict $\widehat{M}_i$ for all selected representative links, and all discrete time-intervals  of  the $i$'th day simultaneously. The choice of the simple network architecture is by design, such that as much of the semantic structure of the temporal condition is encoded into $\mathbf{W}^\textit{emb}$.
Tuning parameters of the embeddings model include the embedding dimensionality, $D$, the pruning factors, $\beta$, $\gamma$, and the size of the \textit{KNN imputation} kernel.

\subsection{Link travel time prediction model}
\label{sec:travel_time_model}
\begin{figure}[ht!]
    \centering
    \includegraphics[scale=1.1]{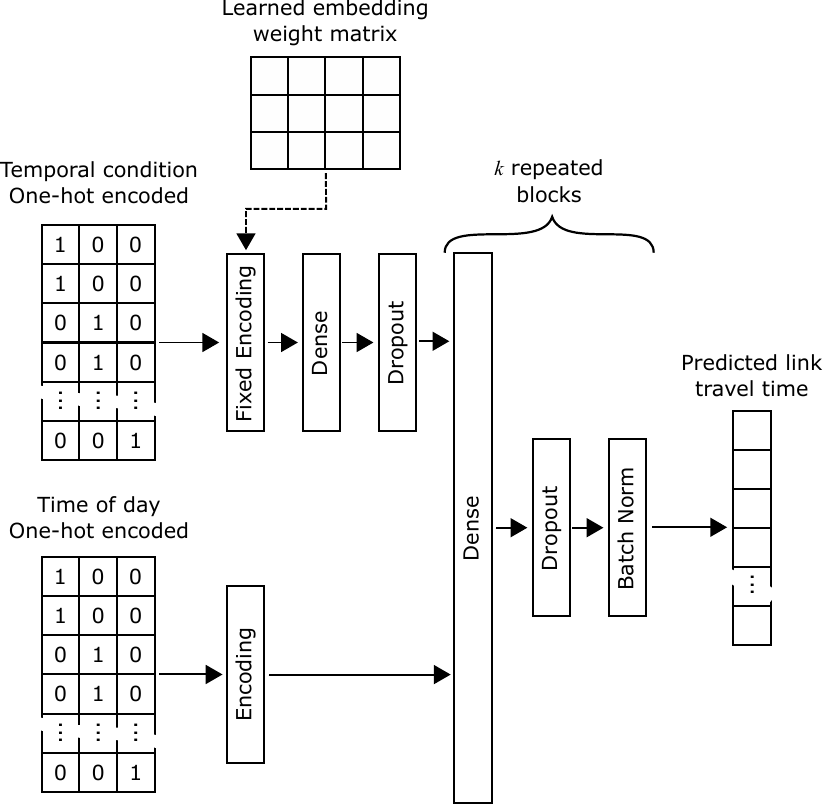}
    \caption{Link travel time prediction model}
    \label{fig:travel_time_model}
\end{figure}

We use the learned embeddings, $\mathbf{W}^\textit{emb}$, as a non-trainable, fixed weight in our link travel time prediction model as shown in Figure~\ref{fig:travel_time_model}. We allow of some transformation of the input using a relatively small and constrained dense layer just on the temporal condition vector representation.

The model also uses the time of day as input, similar to previous research~\cite{Jeong2005,Lin2013}. But we additionally fuse the time of day input with the pre-trained (ajusted) temporal condition vector representation. We could include other inputs to the model, like weather \cite{Wang2018}, traffic condition \cite{Ma2019}, etc. However to keep comparison with respect to the impact of (rare) temporal conditions clear, we refrain from adding additional inputs.
After the fusion between the time of day and the temporal condition representation we allow for a flexible amount of $k$ repeated blocks consisting of a dense layer and bacth normalization. The flexibility is added to allow modelling both simple and complex links, and should just be seen as an additional hyperparameter for the model. During training we additionaly include \textit{dropout} layers \cite{Dropout} to reduce possibility of overfitting our model. We further center and scale the link travel time prior to training using the \textit{mean} and \textit{standard deviation}, and apply the inverse transformation after prediction to return to the original scale. This simply allows for faster and more stable convergence of the network during training.

For training we use the following union of subsets of training data: 1)~All travel time measurements from the $N_\textit{freq}$ \textit{normal} days preceding the start time of the prediction window (i.e.\ days where the \textit{rare condition}-attribute in the calendar data set is \textit{False}.), and 2)~All occurrences of \textit{rare} temporal conditions within $N_\textit{rare}$ days preceding the start time of the prediction window (i.e.\ days where the \textit{rare condition}-attribute in the calendar data set is \textit{True}.). The union of rare and frequent travel times measurements facilitates transfer of information in the network, allowing older travel time measurements for rare temporal conditions to be adjusted by recent travel time measurements for frequent temporal conditions. This is only possible, since we have a coherent semantic structure including both rare and frequent temporal conditions.

\section{Experiments}

In our experiments we use, for the sake of simplicity, 24 equal length discrete time intervals of one hour, and 3 years (1096 days) of travel time measurement data for training the embeddings model. We keep an additional year of data for testing our link travel time models. Thus the shape of $\mathbf{M}^{\textit{ln}}$ for each link, $\textit{ln}$, is $1096 \times 24$.

In the training dataset we have data from a total of $3\,102$ distinct links. We construct $\mathbf{M^\textit{ln}}$ for each of those links and in total we aggregate more than 165 mill. travel time observations from the training data set. To reduce, we prune all links that does not satisfy $\textit{Cov}^\textit{ln} > \beta$, with $\beta = 70\%$. In other words, we will not base the learned semantic representation on links with sparse data, e.g.\ where we are missing data more than $30\%$ of the time in the 3 year training data period. This initially reduced the links considered to $1\,486$. We fill the maximum of $30\%$ missing data points using a \textit{KNN imputation kernel} of size $5$, yielding a complete covered matrix $\mathbf{\widehat{M}}^\textit{ln}$ for each of the $1\,486$ considered links. The overall primary reason for missing data in $\mathbf{M^\textit{ln}}$ for a given link, $\textit{ln}$, is that the link is not scheduled to be serviced in all of the discrete time intervals, e.g.\ no data for night hours.

\begin{figure}[ht!]
    \centering
    \begin{subfigure}[b]{0.48\textwidth}
        \centering
        \includegraphics[width=\textwidth]{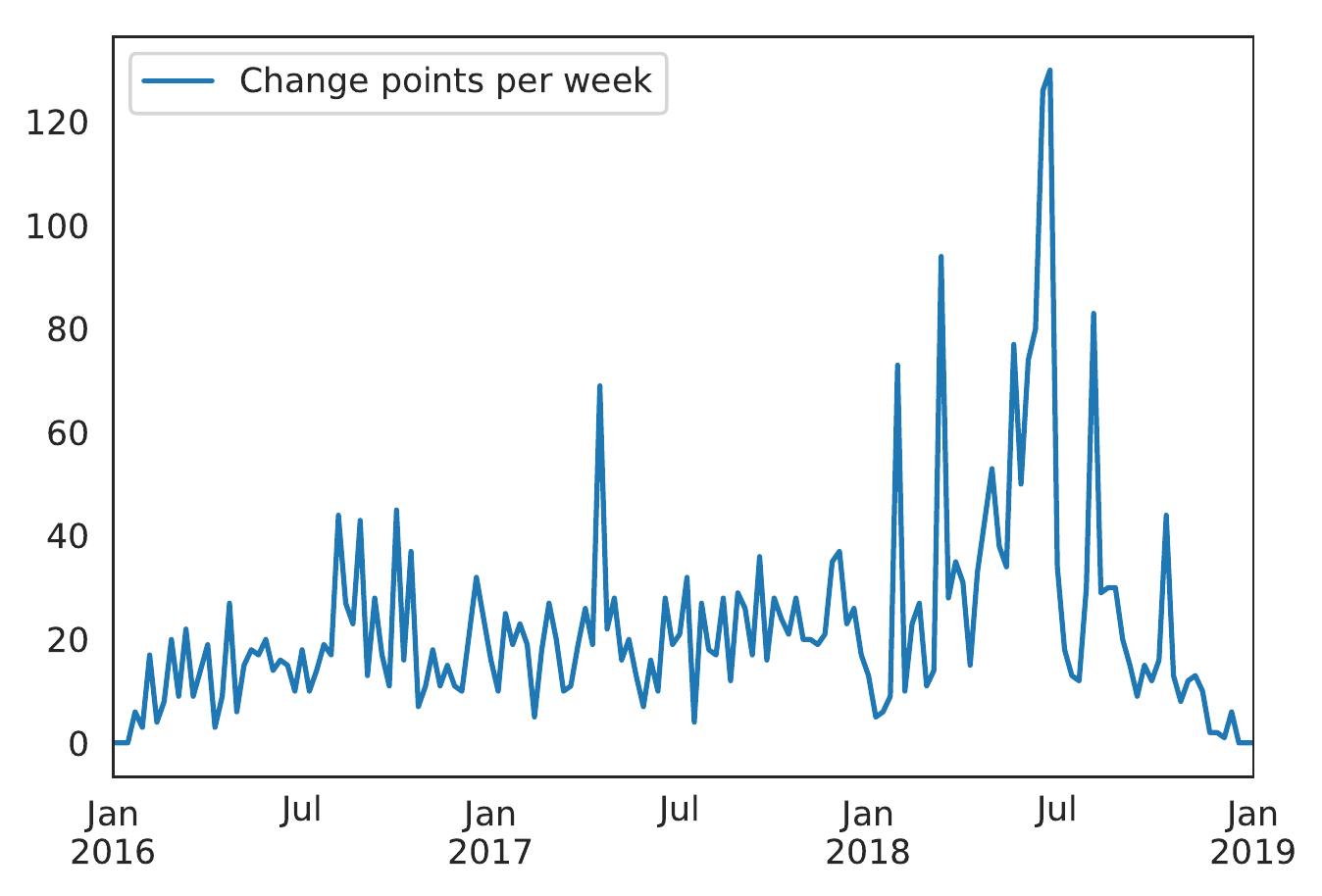}
        \caption{Number of identified change points per week over the three years used for training the embeddings model.}
        \label{fig:change_point_per_week}
     \end{subfigure}
     \hfill
     \begin{subfigure}[b]{0.48\textwidth}
        \centering
        \includegraphics[width=\textwidth]{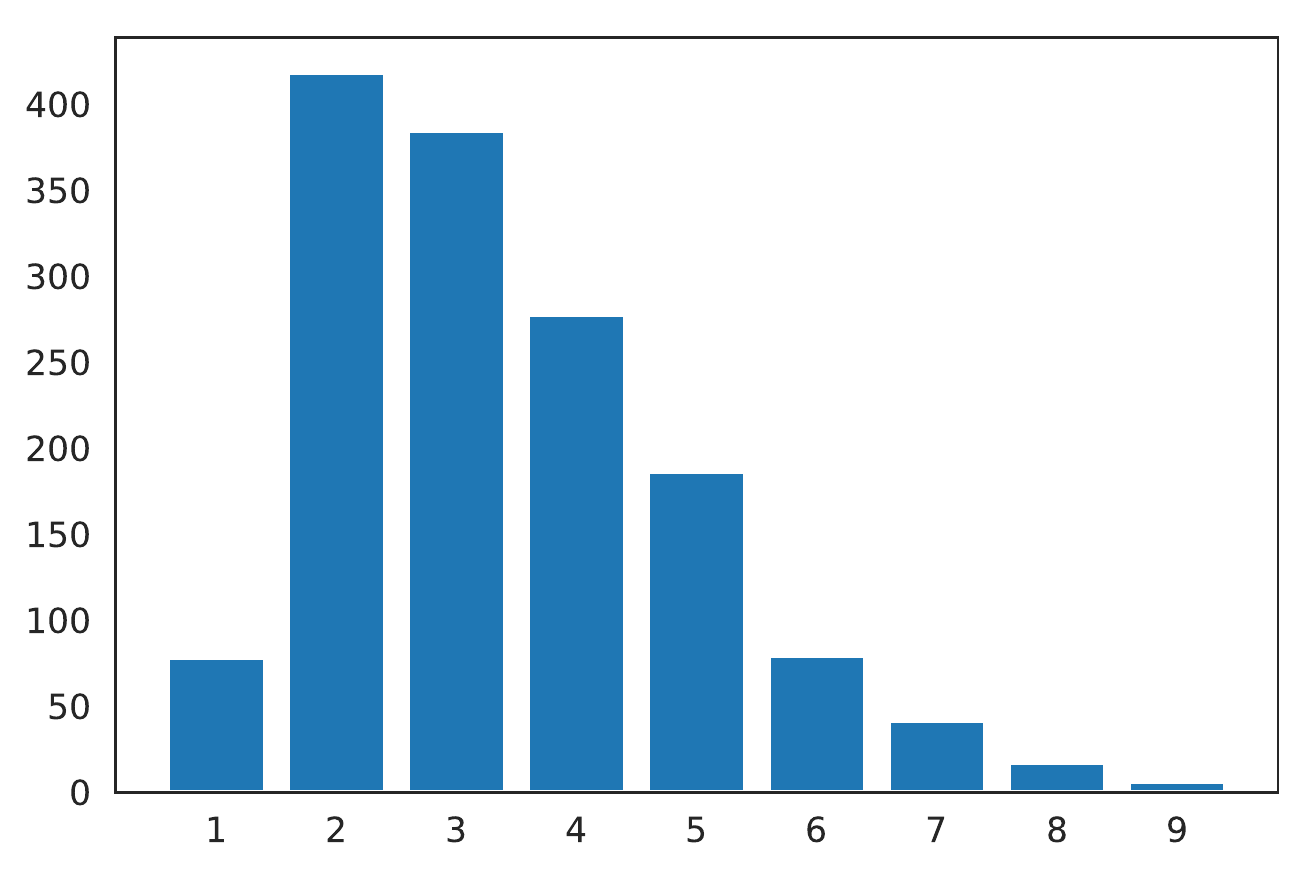}
        \caption{Distribution of number of identified change points.\newline}
        \label{fig:change_point_hist}
     \end{subfigure}
    \caption{Results from running \textit{PELT} change point algorithm on training data.}
    \label{fig:change_point}
\end{figure}
We continue to apply the \textit{PELT} change point algorithm on $\mathbf{\widehat{M}}^\textit{ln}$ for each link, $\textit{ln}$. Figure~\ref{fig:change_point_per_week} shows the occurrence of identified change points over time (summed per week), and Figure~\ref{fig:change_point_hist} the distribution of identified number of change points per link. Notice that all links has at least one identified change point, since the first data point is always considered a change by the \textit{PELT} algorithm implementation used.

Figure~\ref{fig:change_point_per_week} also demonstrates the need for our research, since one can expect a continuously amount of changes. In our experiments we identified approximately 20 changes to link travel times each week. The increased number of change points identified in January to June of the last year of the training set is primarily a result of a slowly rolled-out change to the underlying system recording the link travel time measurements. We have not been able to compensate for such system bias in the data acquisition and preparation process.

\begin{figure}[ht!]
    \centering
    \begin{subfigure}[b]{0.48\textwidth}
        \centering
        \includegraphics[width=\textwidth]{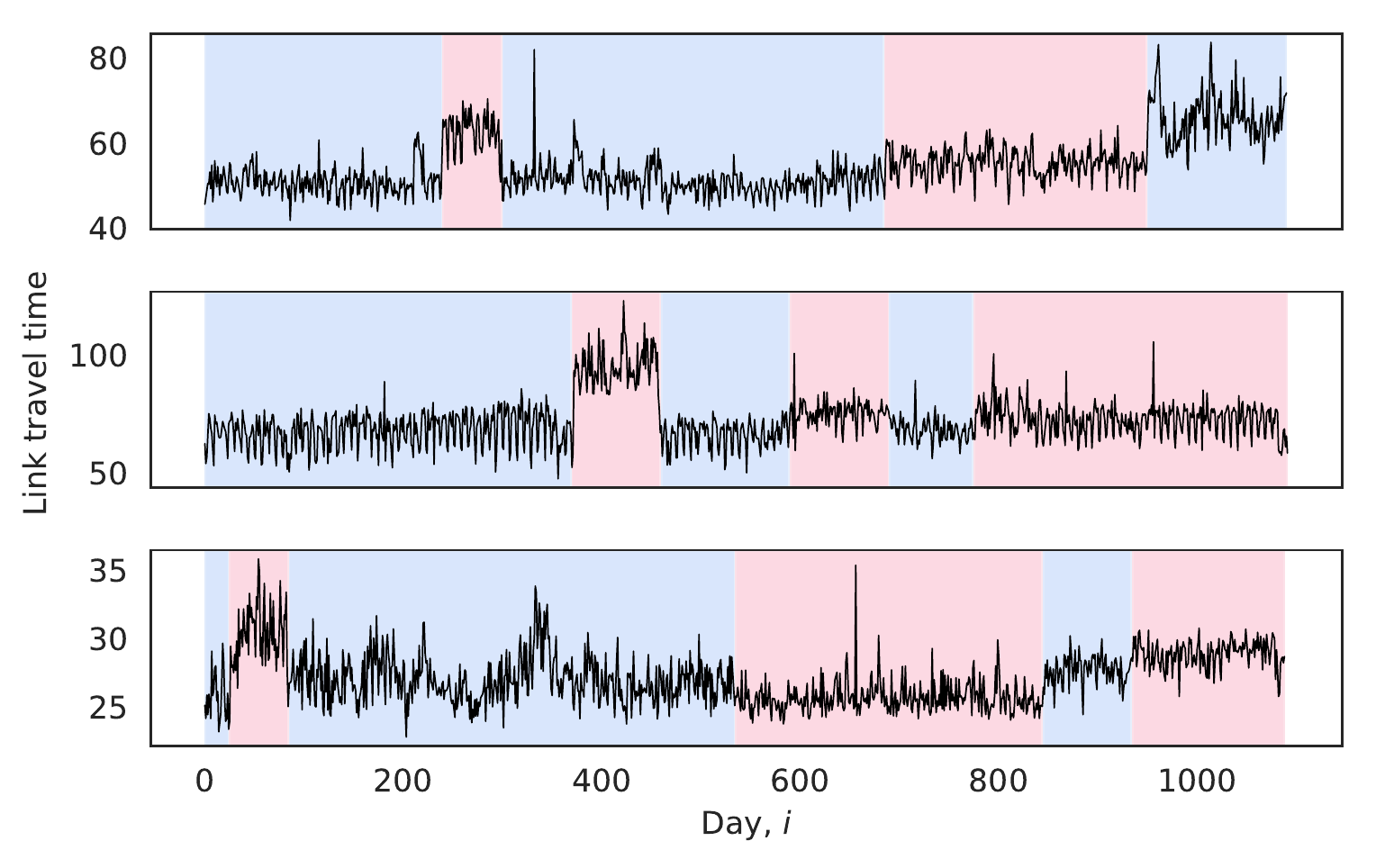}
        \caption{Non-stationary examples, $R^\textit{ln} = 4$}
        \label{fig:non-stationary-examples}
     \end{subfigure}
     \hfill
     \begin{subfigure}[b]{0.48\textwidth}
        \centering
        \includegraphics[width=\textwidth]{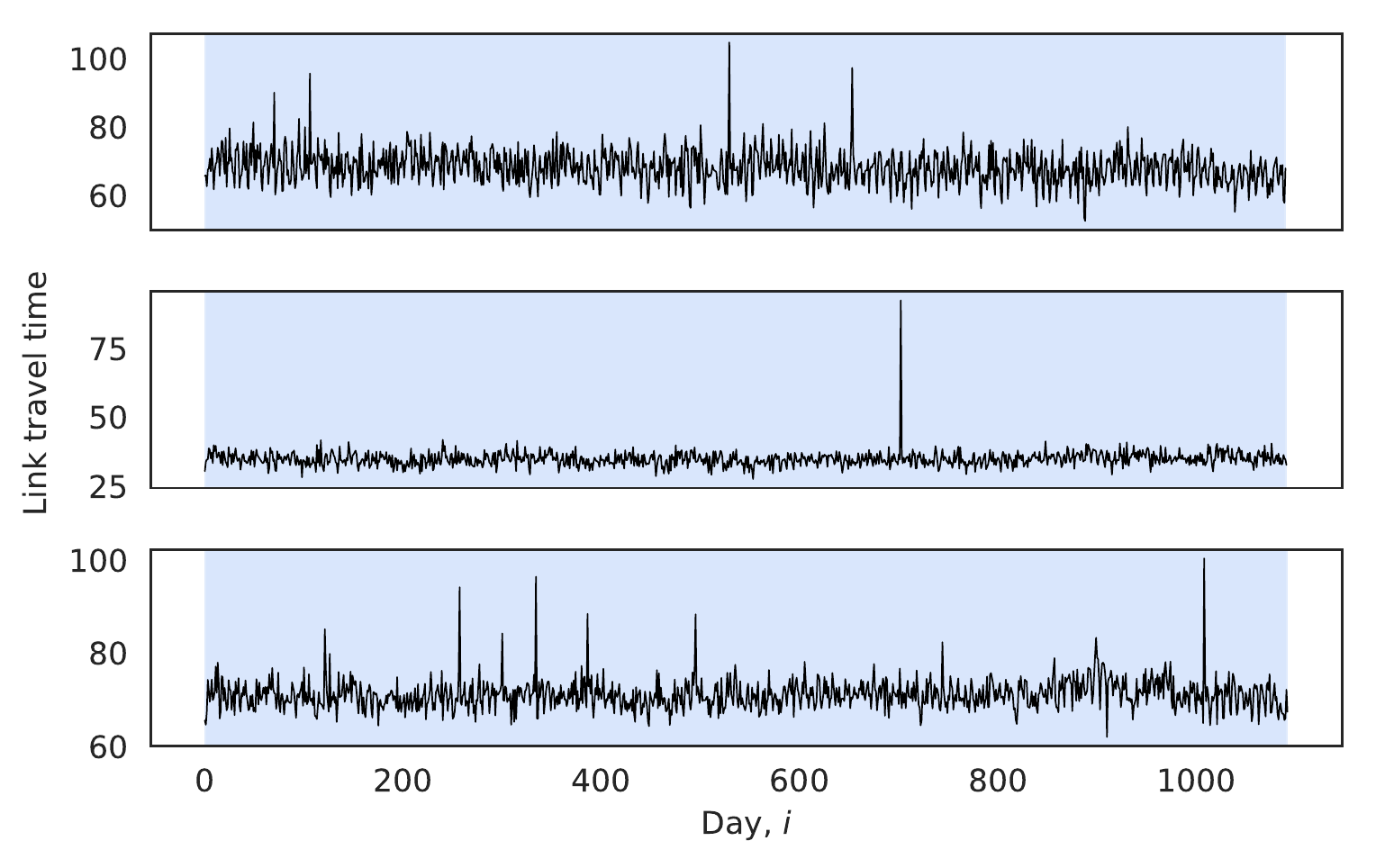}
        \caption{Stationary examples, $R^\textit{ln} = 1$}
        \label{fig:stationary-examples}
     \end{subfigure}
    \caption{Examples illustrating the change points identified by the \textit{PELT}  algorithm.}
    \label{fig:change_point_examples}
\end{figure}
Figure~\ref{fig:change_point_examples} shows three random examples of overall non-stationary links~(a), respectively stationary links~(b) as identified by the \textit{PELT} change point algorithm, where changes in shading indicates an identified change point. 

Based on the distribution of identified change points (i.e.\ Figure~\ref{fig:change_point_hist}), we choose $\gamma = 1$, since we have a fair amount of links (70), which have no additional identified change points.

For the calendar data set (ii) we construct a calendar where official national holidays are marked using their name as the \textit{temporal condition label}. We additionally mark some days based on domain knowledge: adjacent days to holidays, e.g.\ \textit{Between Christmas/New Year Days}, and school vacations periods. Other days are simply marked with their weekday name as \textit{temporal condition label}. In total the calendar data set consists of 20 distinct temporal condition labels. 

\begin{figure*}[ht!]
    \centering
    \includegraphics[width=\textwidth]{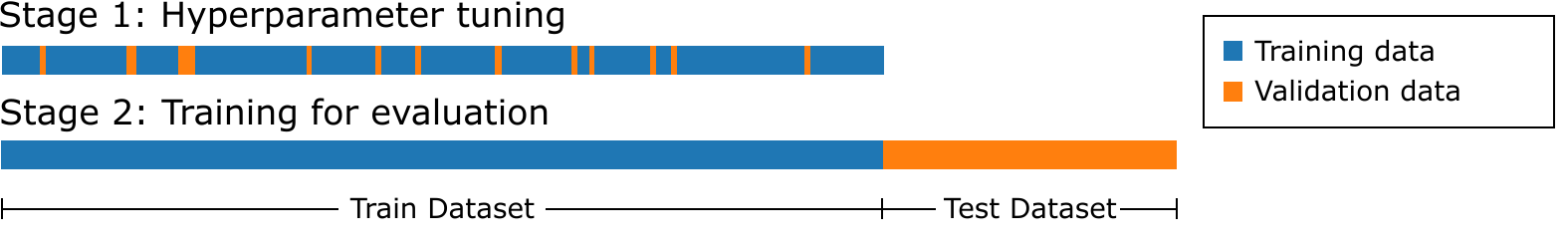}
    \caption{The two-stage process of hyperparameter tuning.}
    \label{fig:tuning_process}
\end{figure*}

We have implemented the embeddings model described in Section~\ref{sec:embeddings_model} in TensorFlow using the Keras API \cite{tensorflow, Keras}. Training and validation is performed using a 20\% randomly sample of the 3 year training data as validation data to tune the hyperparameters. This allowed us to try different numbers of embedding dimensionality, $D$, and different values of $\beta$ and $\gamma$. Once tuned we retrain the embeddings model using the entire training dataset for evaluation on the test dataset. The process of hyperparameter tuning is illustrated in Figure~\ref{fig:tuning_process}.

\begin{figure}[ht!]
    \centering
    \includegraphics[width=3in]{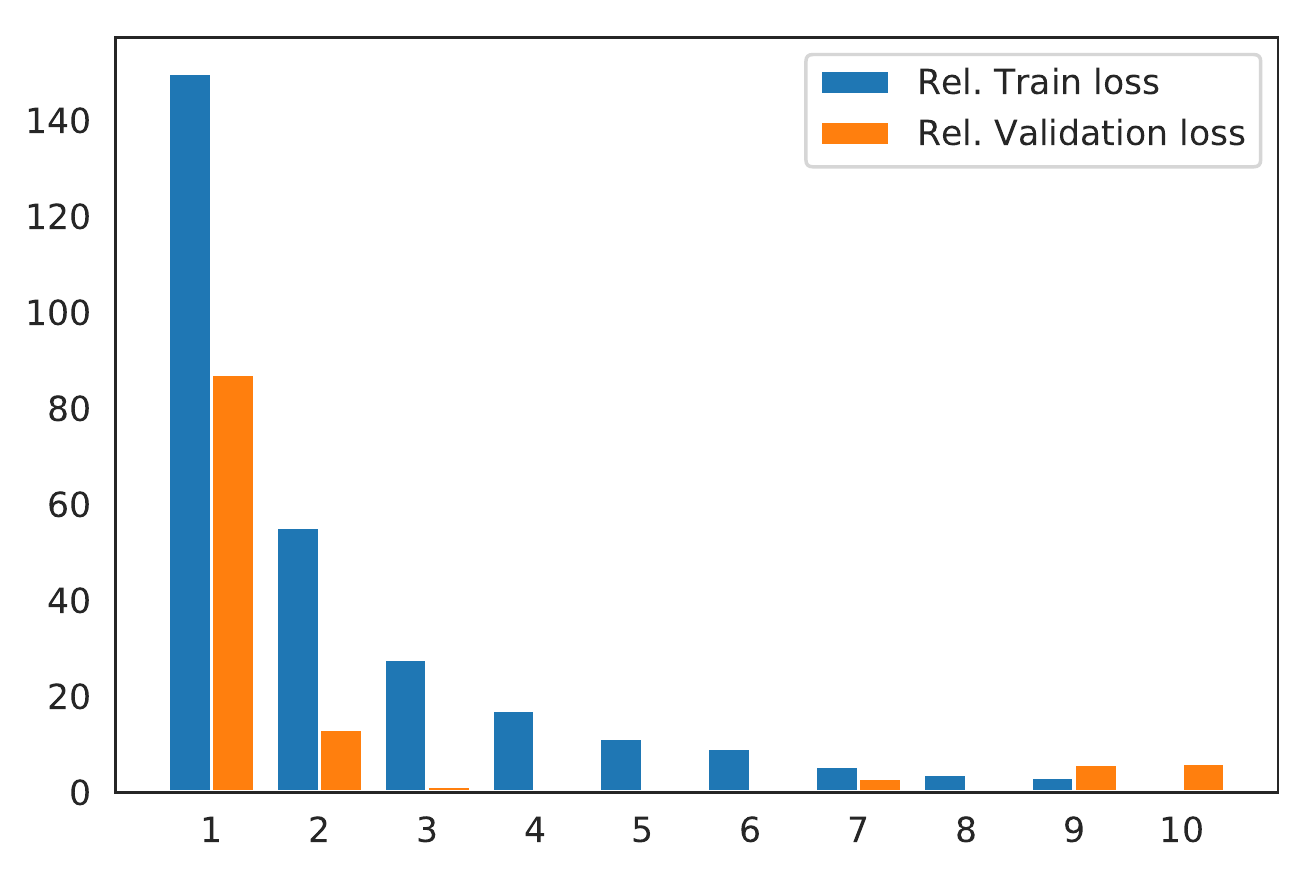}
    \caption{Training of validation loss of embeddings model using different embedding dimensionality, $D$}
    \label{fig:embeddings_dim}
\end{figure}

We see a minimum in validation loss when $D = 4$ as illustrated by Figure~\ref{fig:embeddings_dim}. We see overfitting and in general very little improvement for the train loss with a embeddings dimensionality above 6. Based on this we fix $D = 4$ for our experiment.

\begin{figure*}[ht!]
    \centering
    \includegraphics[width=\textwidth]{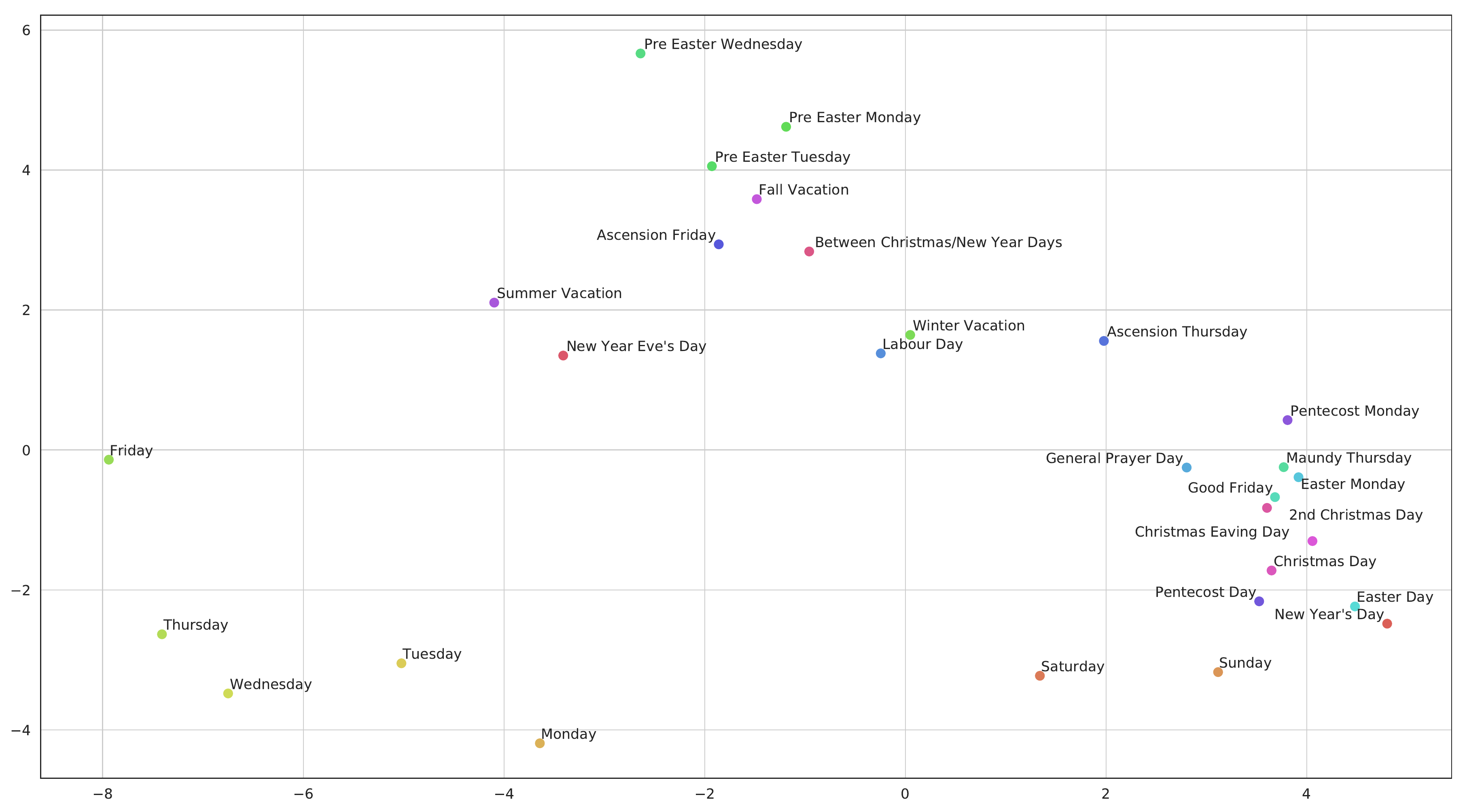}
    \caption{Multidimensional scaling (MDS) of learned day type embeddings.}
    \label{fig:embeddings_mds}
\end{figure*}

To visualize the learned semantic representation encoded into $\mathbf{W}^\textit{emb}$, Multidimensional scaling (MDS) \cite{BorgGroenen2005} is applied. The resulting visual representation is shown in Figure~\ref{fig:embeddings_mds}.
It is worth noticing, that MDS will only preserve pairwise distances. It is though visible that the learned representation have grouped some of the temporal conditions closer together, e.g.: 1) workdays (e.g.\ Monday - Friday), 2) official holidays (e.g.\ Easter Day and New Year's Day), and 3) non-official special days (e.g.\ Pre Easter Tuesday and Ascension Friday). Again, the true learned representations is 4 dimentional vectors, $\textbf{w} \in \mathbb{R}^D$, with $D=4$.

\subsection{Evaluation}
For evaluation of the proposed methodology we seek to evaluate: 1) the ability to learn a semantic representation of temporal (rare) conditions based on a subset of overall homogeneous and representative selection of links, and 2) the ability to utilize and generalize this semantic representation for improved link travel time prediction for non-selected links. To achieve this we evaluate 3 groups each consisting of 10 random sampled links:
\begin{itemize}
    \item \textbf{Embeddings}: This group consists of links from the set of representative links. Performance for links in this group shows the ability to persist information learned from the temporal conditional labeling.
    \item \textbf{Non-selected}: This group consists of links that did not comply with the set criteria, i.e.\ sparse links, $\textit{Cov}^\textit{ln} \leq \beta$, and/or links with heterogeneous travel time time series, $R^\textit{ln} > \gamma$.
    \item \textbf{Unseen}: This group consist of links that are only present in the test data-set, i.e.\ roads/routes that were constructed/created sometime after the 3 year training period, but before our test period cf.\ below. We have no previous observation of any specific temporal condition of the links in this group since the links did not exists at the last occurrence of any temporal condition.
\end{itemize}

To compare our evaluation results of the proposed methodology we also include 3 baselines. The evaluation setup is to predict the travel time of the week of Easter of the test year (the 4th year), consisting of $12\,356$ individual travel time observations for the 30 evaluation links. The size of the training data set depends on the baseline as described below. Again this train and test dataset is on the form illustrated by Table~\ref{tab:data}. For our proposed link travel time prediction model using fixed embeddings weights we use $N_\textit{freq} = 21$ days and $N_\textit{rare} = 365$ days.
\begin{itemize}
    \item \textbf{Historical average with public holidays as Sundays}: Even though this baseline is very simple, it is the currently implemented model for travel time predictions for the Public Transport Authority in the Greater Copenhagen Area. This is arguably a result of public holidays often being timetabled as Sundays in Public Transportation Services. Training data consists of the 21 days preceding the week of Easter. We calculate the hourly average travel time for each link for each weekdays. All public holidays are considered as Sundays.
    \item \textbf{Replicate last year}: Another simple baseline is simply to predict a given temporal condition by replication the last occurrence of it, usually from the previous year. Since we are predicting travel times of individual traversals of links we replicate the closets match with respect to time of day. This is clearly not possible for the \textit{Unseen} test group, since no such previous occurrence exists.
    \item \textbf{Neural network}: Essentially the same as the proposed link travel time prediction model using the fixed embedding, but instead of inputing the temporal condition $\mathbf{c}^\text{oh}_t$, we input the day of week, $\mathbf{dow}^\text{oh}_t$, similary encoded using one-hot encoding. We allow the weight applied to this input to be trainable. Training data consists of the $N$ \textit{normal} days preceding the week of Easter, i.e.\ days where the \textit{rare condition}-attribute in the calendar data set is \textit{False}.
\end{itemize}

Both when training the proposed link travel time prediction model using the fixed learned embeddings, and when training the \textit{neural network baseline} with trainable weights applied to $\mathbf{dow}^\text{oh}_t$, we utilize the Hyperband-algorithm~\cite{Hyperband} for hyperparameter tuning (e.g.\ number of neurons in each layer, number of repeated blocks, $k$, dropout probability, number of epochs, etc.). We apply a similar process for hyperparameter tuning as before and illustrated by Figure~\ref{fig:tuning_process}: We use 20\% random sample of the training data as validation data for choosing hyperparameters. When the hyperparameters has been fixed we retrain using the entire training data set for final evaluation against the test data set.

We have implemented both the temporal conditions model described in Section~\ref{sec:travel_time_model}, and the neural network baseline model described above in TensorFlow using the Keras API \cite{tensorflow, Keras}.

\subsection{Results}
\Cref{tab:rmse} shows the performance results for the three baselines and the temporal conditions model using the \textit{root mean squared error} (RMSE) evaluation score. Similarly \Cref{tab:mae} shows this using the \textit{mean absolute error} (MAE) evaluation score.

For the \textit{embeddings} group, we see replicating last year is yielding the best performance for both RMSE and MAE, and yet it also yields the worst performance for the \textit{non-selected} group. This is not that surprising, since the \textit{embeddings} group specifically consist of overall homogeneous time series, whereas \textit{non-selected} group consist of overall heterogeneous time series. This shows that replicating previous occurrences of rare temporal conditions (i.e.\ Week of Easter travel times) only works when the underlining structure is stable and has not changed since last occurrence. For rare conditions that occur once per year, this becomes unlikely for a significant amount of links. E.g.\ in our experiment we observed an average of 20 links undergoing structural changes per week. Another limitation with replicating previous occurrences is obviously if no such occurrence exists, which is why we cannot evaluate the \textit{replicate last year} baseline for the \textit{unseen group}.

\begin{table}[ht!]
    \centering
\begin{tabular}{lrrr}
&  \multicolumn{3}{c}{Data set} \\
\toprule
Model &  Embeddings &  Non-selected &  Unseen \\
\midrule
 Public holidays as Sundays              &        16.8 &          6.7 &    16.8 \\
Replicate last year &        \textbf{ 9.5} &          9.4 &     - \\
Neural network                  &        13.2 &          5.6 &    13.3 \\
Temporal conditions          &         9.9 &          \textbf{5.0} &    \textbf{10.9} \\
\bottomrule
\end{tabular}
    \caption{Root mean squared error (RMSE)}
    \label{tab:rmse}
\end{table}

\begin{table}[ht!]
    \centering
\begin{tabular}{lrrr}
&  \multicolumn{3}{c}{Data set} \\
\toprule
Model &  Embeddings &  Non-selected &  Unseen \\
                        
\midrule
Public holidays as Sundays               &        12.4 &          5.0 &    12.4 \\
Replicate last year &         \textbf{7.4} &          8.2 &     - \\
Neural network                    &         9.7 &          4.3 &    10.1 \\
Temporal conditions          &         7.9 &          \textbf{3.9} &     \textbf{8.9} \\
\bottomrule
\end{tabular}
    \caption{Mean absolute error (MAE)}
    \label{tab:mae}
\end{table}

We also notice from \Cref{tab:rmse,tab:mae} that our proposed methodology performs only 4\%, respectively 7\% worse than the best for the \textit{embeddings} group, while it is the best model for both the \textit{non-selected} and \textit{unseen} groups. For the \textit{non-selected} group it performs 11\%, respectively 9\% better than the neural network baseline. For the \textit{unseen} group it performs 18\%, respectively 12\% better than the neural network baseline.


Lastly it is worth noticing that predicting public holidays as Sundays are not a valid approach in general as it yields the worst results for the \textit{embeddings} and \textit{unseen}. This supports our initial point, that rare temporal conditions does not constitute a single distribution of neither travel demand nor travel time.


\section{Conclusion}
We have presented a vector-space model for learning the semantic representation of both rare and frequent temporal conditions for travel time. Both frequent and rare conditions are part of a single coherent semantic structure, allowing links with limited historic data for rare conditions to benefit from historic data of frequent conditions. The learning is done using a subset of dense link time series in order to avoid exaggerated use of imputation, and overall homogeneous time series of link travel time. The latter ensures that we indeed learn the impact of the temporal conditions, and not other structural changes made over time, e.g.\ road works, introduced traffic calming initiatives, changes to intersection signal programs, etc.

We applied to method to bus travel times from the Greater Copenhagen Area. Our main findings from our experiment are:
\begin{itemize}
    \item \textbf{The learned embeddings generalize}: Our evaluation and results shows that the learning the semantic representation generalizes to other links not used for training the semantic representation (\textit{non-selected} group), and even to new links that has no history of rare temporal conditions (\textit{unseen} group).
    \item \textbf{Holidays are not just Sundays}: Many holidays have completely different travel patterns than a regular Sunday, and therefore modeling them as such causes significant errors. Likewise, including general holidays in the dataset for estimating Sundays will result in decreased performance for this task.
    \item \textbf{Repeating previous years holiday} only works for links with overall homogeneous travel time, and fails with high errors for links that has been subject to changes in the underlining structures. The approach is not a viable option for new links added to the network since last occurrence of a given temporal condition, or links with limited historical data in general. 
\end{itemize}


\subsection{Future work}
We have showed results for encoding rare temporal conditions for the bus travel time problem. We would like to test the method on general travel demand and travel time for several modes: \textit{demand prediction}, \textit{logistics planning}, and \textit{general planning}. Furthermore we would like to investigate whether the proposed method can by applied to other rare conditions, such as rare weather conditions.

\bibliographystyle{IEEEtran}
\bibliography{local}
\newpage

\end{document}